\documentclass[journal]{IEEEtran}
\usepackage{times}
\usepackage{epsfig}
\usepackage{graphicx}
\usepackage{amsmath}
\usepackage{amssymb}
\usepackage{subfigure} 
\usepackage[dvipsnames, svgnames, x11names]{xcolor}
\usepackage[ruled,vlined]{algorithm2e}
\usepackage{algorithmic}
\usepackage{multirow}
\usepackage{stfloats}
\usepackage{svg}
\usepackage{float}
\usepackage{threeparttable}
\usepackage{multirow}

\usepackage{cite}
\begin{document}
%
\title{A Dynamic GCN with Cross-Representation Distillation for Event-Based Learning}
%
%
%

\author{Yongjian~Deng,\thanks{Yongjian Deng is with the College of Computer Science, Beijing University
of Technology, Beijing 100124, China. {\tt\small yjdeng@bjut.edu.cn}.}
        Hao~Chen,~\IEEEmembership{}\thanks{Hao Chen is with School of Computer Science and Engineering, Southeast University, and Key Lab of Computer Network and Information Integration (Southeast University), Ministry of Education, Nanjing 211189, China. {\tt\small haochen593@gmail.com}.}
        Bochen~Xie,~\IEEEmembership{Member,~IEEE,}\thanks{Bochen Xie and Youfu Li are with Department of Mechanical Engineering,
        City University of Hong Kong, Kowloon, Hong Kong SAR.
        {\tt\small $\{$ boxie4-c@my., meyfli@$\}$cityu.edu.hk}.}\\
        Hai~Liu,~\IEEEmembership{Senior Member,~IEEE}\thanks{Hai Liu is with the National Engineering Research Center for E-Learning, Central China Normal University, Wuhan 430079, China {\tt\small hailiu0204@ccnu.edu.cn.}}
        and~Youfu~Li,~\IEEEmembership{Fellow,~IEEE}}
\markboth{}{Deng \MakeLowercase{\textit{et al.}}: A Dynamic GCN with Cross-Representation Distillation for Event-Based Learning}
%



\maketitle

\begin{abstract}
Recent advances in event-based research prioritize sparsity and temporal precision. Approaches using dense frame-based representations processed via well-pretrained CNNs are being replaced by the use of sparse point-based representations learned through graph CNNs (GCN). Yet, the efficacy of these graph methods is far behind their frame-based counterparts with two limitations. ($i$) Biased graph construction without carefully integrating variant attributes ($i.e.$, semantics, spatial and temporal cues) for each vertex, leading to imprecise graph representation. ($ii$) Deficient learning because of the lack of well-pretrained models available. Here we solve the first problem by proposing a new event-based GCN (EDGCN), with a dynamic aggregation module to integrate all attributes of vertices adaptively. To address the second problem, we introduce a novel learning framework called cross-representation distillation (CRD), which leverages the dense representation of events as a cross-representation auxiliary to provide additional supervision and prior knowledge for the event graph. This frame-to-graph distillation allows us to benefit from the large-scale priors provided by CNNs while still retaining the advantages of graph-based models. Extensive experiments show our model and learning framework are effective and generalize well across multiple vision tasks.\footnote{\textbf{Core components of our codes are submitted with supplementary material and will be made publicly available upon acceptance}.}.
\end{abstract}


%
\IEEEpeerreviewmaketitle

\section{Introduction}



\begin{figure}[t]
  \centering
\includegraphics[width=1\linewidth]{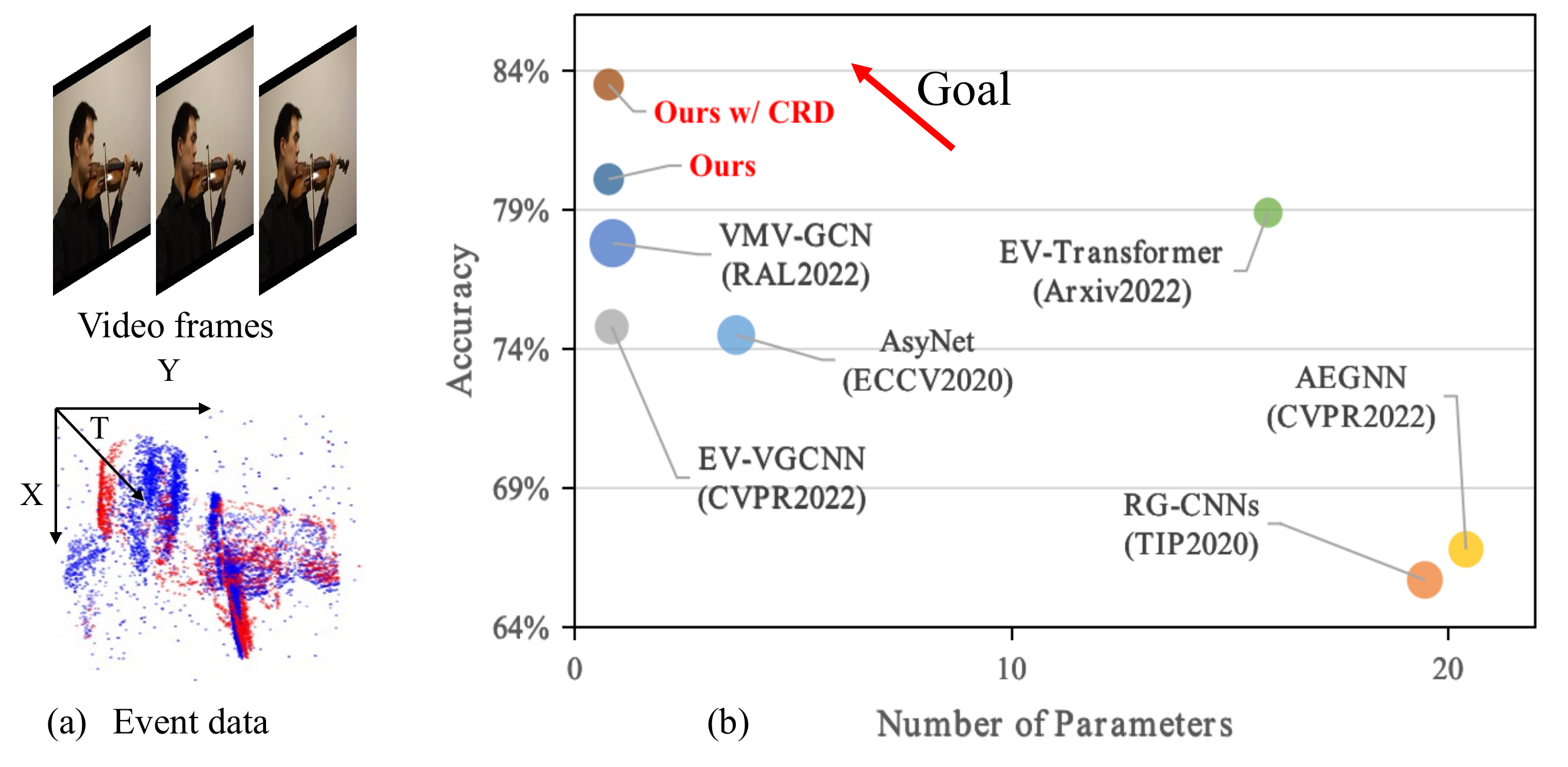}
\caption{(a) Visual comparison between outputs from traditional cameras and event cameras. (b) Recognition accuracy vs model complexity (\#Params) of our approach (EDGCN w/o and w/ CRD) on the N-Caltech dataset. The circle areas are proportional to the computational complexity (FLOPs).}
\label{fig1}
\end{figure}

Each pixel of an event camera can report an event independently when detecting logarithmic brightness changes.\footnote{We place an illustration of the event-based working principle in the supplementary.} This unique working principle provides event cameras with high dynamic range, low power consumption, and high temporal resolution while maintaining data non-redundancy. As a result, event cameras have great potential for use in mobile and wearable devices\footnote{A survey about event-based applications can be found in \cite{gallego2019event}}. However, the effective representation and learning of this special data format (as shown in Figure \ref{fig1}) is still a topic of ongoing research.

Event data representation methods can be divided into two categories: \textbf{frame-based methods} \cite{gehrig2019end, Cannici_2020_ECCV, messikommer2020event, mvfnet, Tore2022pami, ev2vid} that sacrifice data sparsity and motion precision to make event data compatible with pre-trained CNNs, and \textbf{point-based methods} that protect the sparsity and temporal structure of event streams. Point-based methods include PointNet-like architectures \cite{qi2017pointnet, wang_2018_wacv, Sekikawa_2019_CVPR} and event-based graphs \cite{Graph-based, Li21iccv, Schaefer22cvpr, Deng_2022_CVPR, Xie_2022_RAL, li2022event} are designed for aggregating local continuous semantics. However, due to discrepancies among different event attributes ($i.e.$, spatial, temporal coordinates, and semantics) and limited labeled data, existing point-based methods suffer from biased graph representation or inadequate training.



To fully leverage the inherent advantages of event cameras, we focus on address two key problems of current point-based solutions: (1) \textit{how to deal with diverse attributes (semantic, space distance, and temporal cues) to determine vertices’ neighbors}. (2) \textit{How to facilitate event graph learning without relying on additional data}.

\begin{figure}[]
  \centering
\includegraphics[width=1\linewidth]{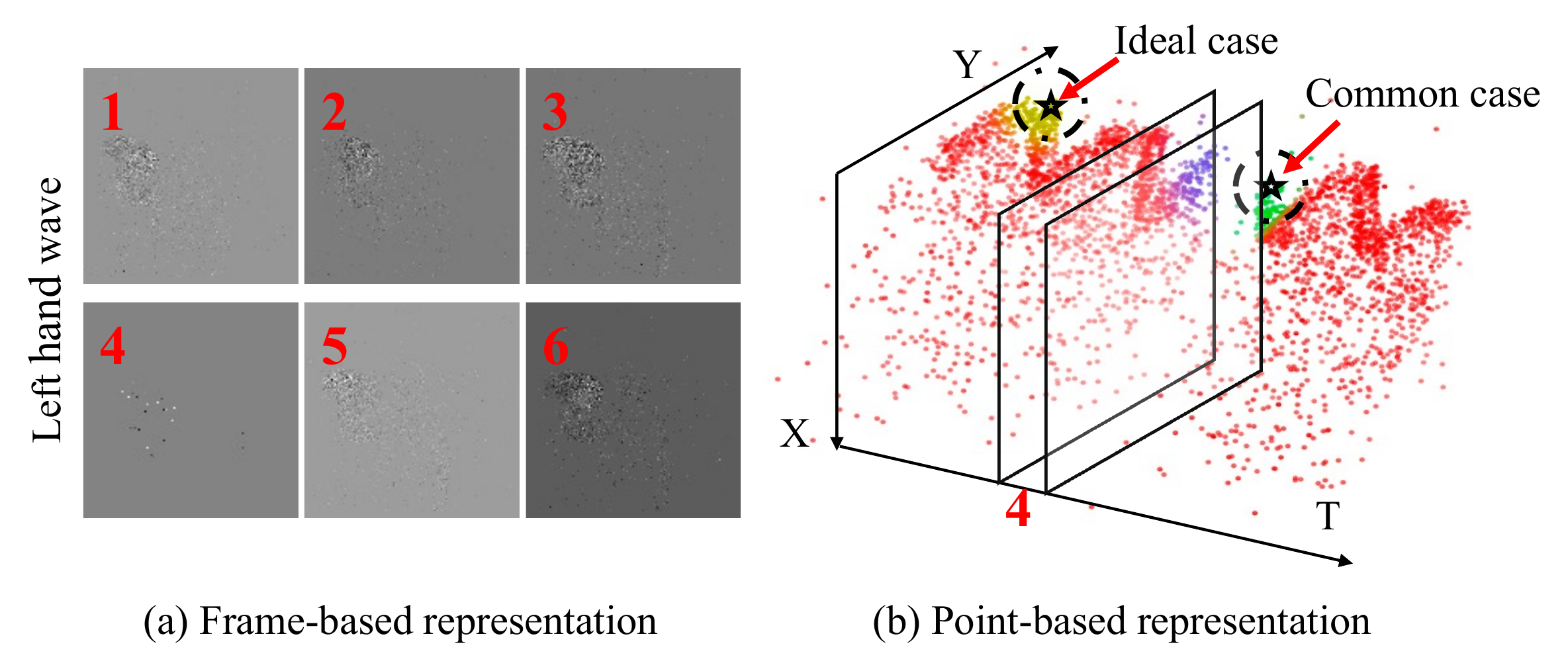}
\caption{Event-based representations for left-hand waving action. (a) Images integrated by events with different time intervals. Event stream  (b) breaks in the fourth interval due to motion stagnation. Coordinate-defined neighborhoods \cite{Graph-based, Deng_2022_CVPR, Schaefer22cvpr, Xie_2022_RAL} can find temporally and semantically related neighbors  for vertices ($\star$) in ideal continuous cases. However, motion stagnation and occlusion problems, common cases in practice, make it hard to find highly related neighbors based on coordinates only ($e.g.$, purple vertex before motion stagnation). Thus, it is necessary to incorporate semantic information into neighborhood definition.}
\label{fig: neighbor_finding}
\end{figure}

Properly handling the neighborhood relations of vertices is key to achieving accurate event graph representations. Although previous works have achieved much success in this direction, they usually follow traditional 3D vision methods and try to define neighborhoods by better handling spatio-temporal coordinates, such as unifying the value ranges of spatial-temporal coordinates \cite{Graph-based, Deng_2022_CVPR, Li21iccv, Schaefer22cvpr} (Figure \ref{fig: neighbor_comparison}.(b)) or dynamically updating coordinates \cite{Xie_2022_RAL, 9181247} (Figure \ref{fig: neighbor_comparison}.(c)). Figure \ref{fig: neighbor_finding} shows that these methods have difficulty in effectively modeling discontinuous event streams caused by motion stagnation or occlusion. To this end, we introduce a simple but effective graph construction strategy that define vertex neighborhoods considering all attributes ($i.e.$, semantic, space distance, and temporal cues) in a learnable manner (Figure \ref{fig: neighbor_comparison}.(d)). To learn from the proposed graph, we introduce an Event-based Dynamic Aggregation Layer (EDAL) that has multi-attribute joint learning branches for defining neighborhood integration strategies. We also include a rebalanced design that uses only coordinates for vertices' attentive aggregation, allowing the model to better judge the motion state of vertices and verify its effectiveness through experiments (Table \ref{tab:EDAL_choice}).

To enhance the learning of event-based graph models, we take full use of the original event streams and map them to an additional frame-based representation branch, and borrow the valuable prior knowledge in well-pretrained CNNs to facilitate the event graph model. We propose a cross-representation (frame-to-graph) distillation framework with a hybrid distillation structure, combining an intermediate feature-level contrastive loss \cite{chen2020simple} and an inference-level distillation loss \cite{hinton2015distilling, Romero15-iclr}. This framework transfers multi-level semantics and handles frame-graph discrepancies across layers. Compared to previous frame-to-frame transfer learning methods, \textit{our method uses only event signals without extra data, while holding better learning and graph model advantages.} Our framework also shows better generalization despite the large representation discrepancy.


The main contributions are summarized as follows: 
\par(1) We newly introduce a graph construction strategy with customized learning model for event data, where the cross-vertex dependency is determined by the joint representation from all attributes of vertices in an unbiased and dynamic way.
\par(2) It is the first cross-representation distillation (frame-to-graph distillation) work for event data with no extra data required. We carefully analyze the variant discrepancy between frame and graph representation in different layers and use corresponding constraints to distill different layers.
\par(3) Overcoming event-graph construction and training bottlenecks, leading to significant accuracy gains with extremely low model and computation complexity.
\par(4) Extensive experiments validate the efficacy of our proposed event graph and learning strategy on various downstream tasks, verifying the high generalization ability of our model.

\begin{figure*}[]
\centering
\includegraphics[width=1\linewidth]{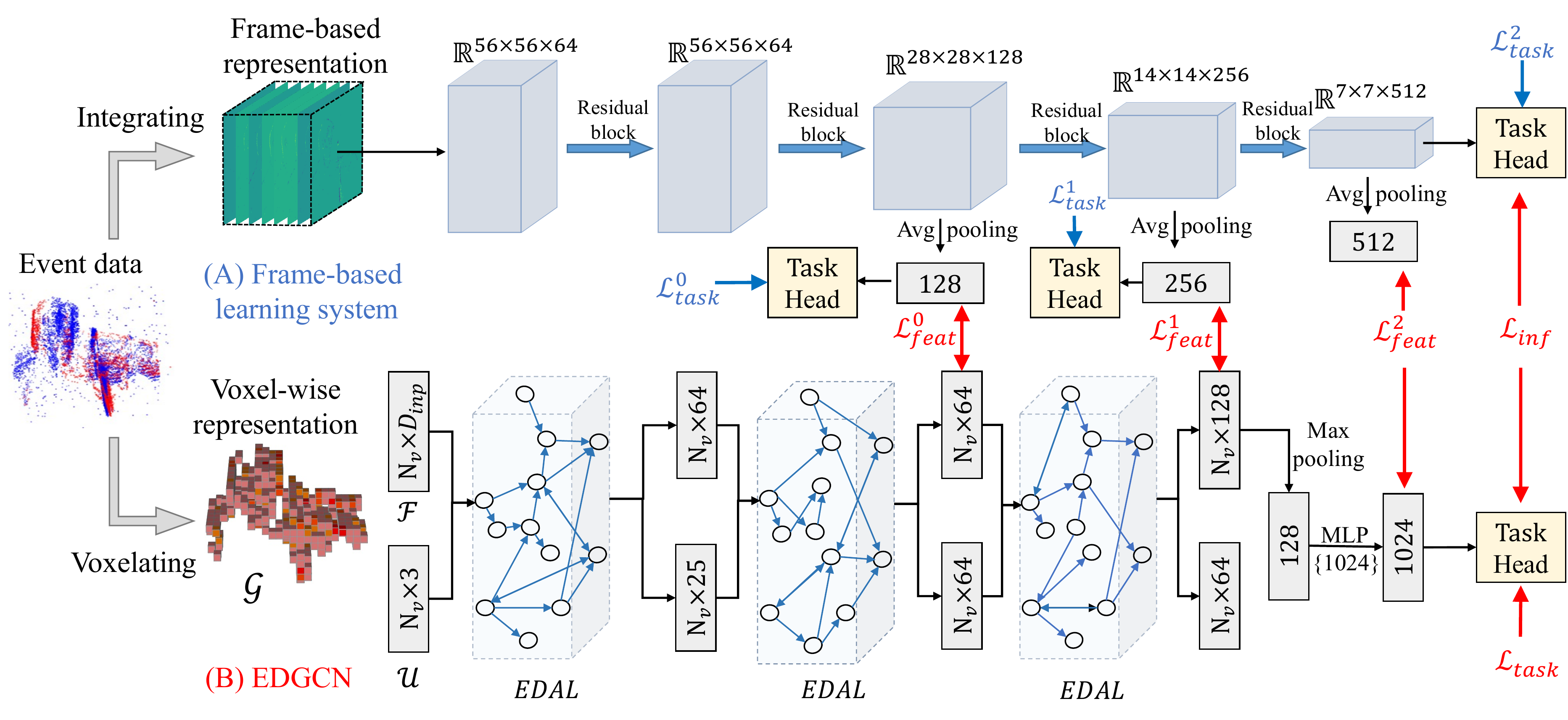}
\caption{The pipeline of our proposed method takes two different representations derived from the same input as input for the frame-based model (A) and the point-based model (B), i.e., EDGCN. A ResNet-like structure is chosen to represent the frame-based model. “MLP” stands for multi-layer perceptron, with numbers in brackets indicating layer sizes. The function of the graph learning module EDAL and losses are depicted in \ref{subsec:EDAL} and \ref{subsec:CRLS}, respectively. Losses in blue are used for frame-based model training, while those in red are imposed when optimizing the EDGCN with CRD. \textit{Only EDGCN is used at the inference stage.}}
\label{pipeline}
\end{figure*}

\begin{figure}[]
  \centering
\includegraphics[width=1\linewidth]{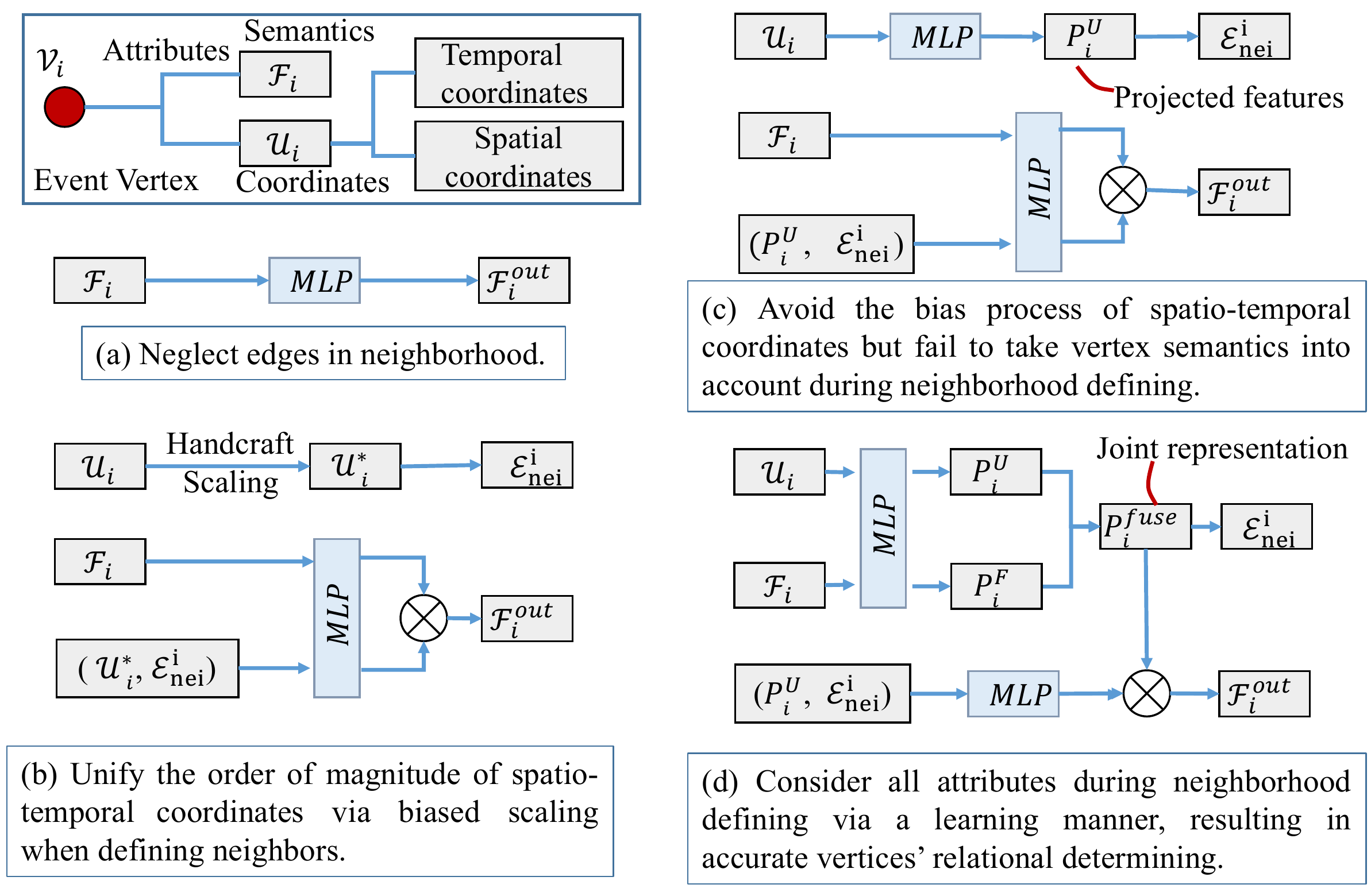}
\caption{Comparison of graph learning manner in representative works. Clear to see that one of the major differences between event-based graph learning is how to define neighborhood ($\mathcal{E}_{nei}$) of a vertex ($\mathcal{V}$) according to various attributes ($\mathcal{F}$ \& $\mathcal{U}$), which may the reason of their performance discrepancy. $\otimes$: Attentive aggregation.}
\label{fig: neighbor_comparison}
\end{figure}

\section{Related Work}\label{sec:relwork}
\noindent{\textbf{Event-based learning}}\indent
Event-based processing can be done using either frame-based or point-based methods. Frame-based methods integrate events into dense representations and adapt these representations to CNNs for further processing, achieving high performance on various vision tasks, $e.g.$, event-based recognition \cite{gehrig2019end, Cannici_2020_ECCV, messikommer2020event, mvfnet, Tore2022pami}, video reconstruction \cite{ev2vid, zhang2021event, Zhu_2022_CVPR}, optical flow estimation \cite{zihao2018unsupervised, hagenaars2021self, Hu_2022_CVPR}. However, these methods sacrifice the sparsity and temporal precision of event data \cite{mitrokhin2020learning, Schaefer22cvpr}, resulting in redundant computation and high model complexity. Point-based methods exploit the sparsity of events to design recognition models with low model complexity and fast inference speed, making them popular. Spiking neural networks (SNNs) \cite{7010933, sironi2018hats} were the first applied to event data due to their sparse and asynchronous nature, but training classic SNNs is intractable. Recent methods combine SNNs with CNNs to mitigate this issue, yet the introduced additional computations weaken the superiority of SNNs. Inspired by the 3D vision, PointNet-like models \cite{wang_2018_wacv, Sekikawa_2019_CVPR} have also been developed for event data, but these structures are limited in aggregating local features adaptively (Figure \ref{fig: neighbor_comparison}.(a)).

Graph-based approaches \cite{mitrokhin2020learning, Graph-based, Li21iccv, Schaefer22cvpr, 9181247} have been proposed to address the problem of heuristic neighborhood definition with respect to the properties of events. Compared to frame-based methods, graph-based methods are more lightweight and show better prospects on various downstream tasks and real-life application scenarios. Recent research has proposed novel representations \cite{Deng_2022_CVPR, Xie_2022_RAL} or investigated spatial proximity of events \cite{li2022event} to close the performance gap to frame-based methods. Yet, defining neighborhoods by considering all vertices’ attributes without bias and facilitating graph model learning with limited labeled data remain unstudied. In this work, we propose a dynamic joint representation learning approach (EDGCN) with a cross-representation distillation training framework (CRD) to address both problems.

\noindent{\textbf{Transfer learning on event data}}\indent
Many transfer learning studies have been proposed to mitigate the difficulties in training event-based models. For instance, \cite{8946715, wang2020eventsr, tulyakov2022time} converts event data to RGB images for better adaption with pretrained CNNs. However, these methods usually suffer from additional computation cost during modality transferring. \cite{zanardi2019cross, Hu_2020_Graft, evkdnet, sun2022ess, messikommer2022bridging} adopt cross-modality transfer schemes by introducing extra supervision to facilitate the event-based learning, while additional visual modalities are not always available in practice. Some other works \cite{Gehrig_2020_CVPR, rebecq2018esim} try to supplement event-based datasets by simulating event data from traditional videos, while artifacts are inevitably introduced. Instead, our proposed cross-representation distillation scheme leverages the transferable knowledge contained in various deep models trained using different representations of event data, where no extra data are required. This approach has a wider range of applications and advances the learning of the event graph branch through a frame-to-graph distillation loss.

\section{Approach}\label{sec:approach}
In this work, we devise a novel event-based dynamic graph CNN (EDGCN) and a cross-representation distillation strategy (CRD) to boost its learning sufficiency further. The pipeline of our proposed method is illustrated in \ref{pipeline}, which contains the following key steps: (1) Representation construction of events for both the graph model and traditional CNNs. (2) Gradual aggregation of contexts using successive event-based dynamic aggregation layers (EDALs). (3) Parallel to the event graph branch, we map the original events to a frame-based branch as the teacher to promote EDGCN branch learning via the proposed cross-representation distillation framework. (4) The EDGCN is appended with different inference heads for various tasks. The following sections will clarify detailed designs and implementations in these steps. 

\subsection{Event-based representation} \label{subsec:evrepre}

\noindent{\textbf{Event data}}\indent
Each event $e_i$ holds three properties: the occurred location ($(x_i, y_i)$), the triggered timestamp ($t_i$), and the polarity ($p_i \in \begin{Bmatrix}-1, 1\end{Bmatrix}$). Particularly, the positive $p$ denotes the brightness increase and vice versa.

\noindent{\textbf{Voxel-wise representation}}\indent
In this work, we directly adopt the voxel-wise representation \cite{Deng_2022_CVPR} of events as our input. In specific, event streams ($\begin{Bmatrix}e_i\end{Bmatrix}_N = \begin{Bmatrix}x_i, y_i, t_i, p_i\end{Bmatrix}_N$) is firstly partitioned into voxels with the voxel size ($v_x, v_y, v_t$). Then, $N_v$ voxels containing the largest number of events are reserved as vertices ($\mathcal{V}$) of the event-based graph ($\mathcal{G}$). Here, we denote the left-upper location of a vertex ($\mathcal{V}_i$) as its coordinate attribute ($\mathcal{U}_i = (x^v_i, y^v_i, t^v_i)$). Finally, the semantics ($\mathcal{F}_i \in \mathbb{R}^{D_{inp}}$) of the $i$-th vertex ($\mathcal{V}_i$) is obtained through event-wise integration formulated by the function $\Omega$, where $D_{inp} = v_xv_y$. \textit{More details about the integration function $\Omega$ can be found in the supplementary material.} 


\subsection{EDGCN}
We address the challenge of modeling dependencies and assigning well-matched contributions of three attributes (spatial position, triggered time, and local semantics) in measuring cross-vertex edges for constructing an event-based graph by using dynamic aggregation layers. Our approach includes neighborhood definition, attentive aggregation, and coordinate attribute update. The dynamically updated graph improves neighborhood definition accuracy and feature aggregation efficacy by continuously refining vertex attributes at each layer.

\begin{figure}[]
\centering
\includegraphics[width=0.9\linewidth]{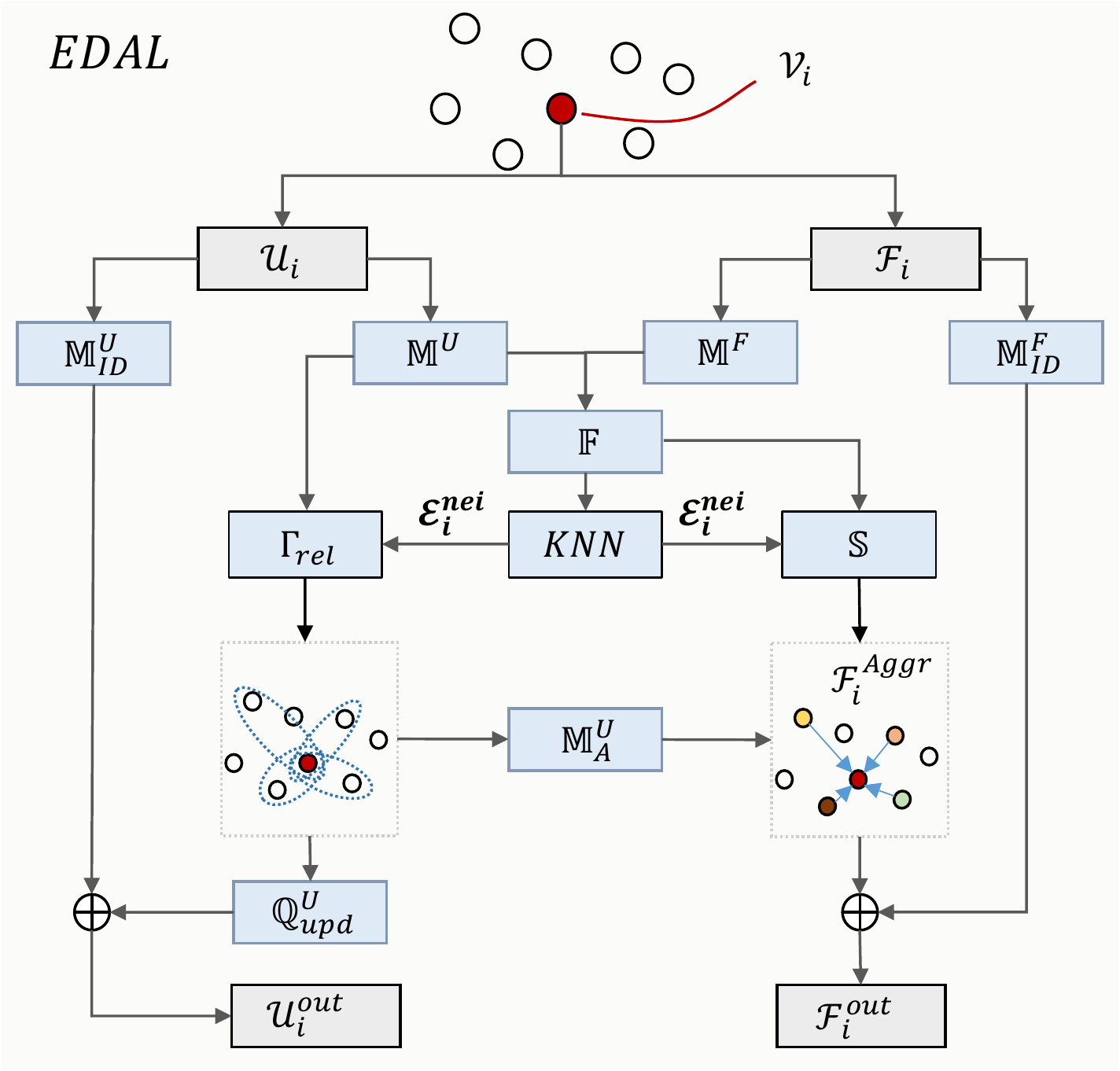}
\caption{An illustration of how the an EDAL aggregates features for vertex $\mathcal{V}_i$ from its neighbors. The function of each notation shown in the figure is detailed in \ref{subsec:EDAL}. $\bigoplus$: element-wise addition.}
\label{EDAL}
\end{figure}

\subsubsection{Dynamic aggregation layer (EDAL)} \label{subsec:EDAL}
The main component of our EDGCN, namely EDAL, is schematized in \ref{EDAL}. Its workflow is detailed successively as follows.

\noindent\textbf{Neighborhood definition}\indent
We suppose that the $i$-th vertex $\mathcal{V}_i$ is an input vertex to an EDAL with coordinate and semantic attributes $Attr(\mathcal{V}_i):(\mathcal{U}_i \in \mathbb{R}^{{D^{in}_u}}, {\mathcal{F}}_i \in \mathbb{R}^{ {D^{in}_f}})$. The goal of EDAL is to define neighborhood space for each vertex by considering all its attributes and aggregating attributes from its neighbors attentively. Considering the large discrepancy in spatial position, triggered time, and features of local semantics, we argue that projecting the $Attr(\mathcal{V}_i)$ to a unified feature space with \ref{projection} is in demand.
\begin{equation}
\begin{aligned}
{\mathcal{P}^F_i} = {\mathbb{M}^F}  ({\mathcal{F}_i}), \ \ \ 
{\mathcal{P}^U_i} = {\mathbb{M}^U}  ({\mathcal{U}_i}),
\end{aligned}
\label{projection}
\end{equation}
where $\mathbb{M}^F$ and $\mathbb{M}^U$ are \textit{MLP}s for feature projection. The obtained representations $\mathcal{P}^F_i$ and $\mathcal{P}^U_i$ $ \in \mathbb{R}^{{D^{in}_f}}$ in the same feature space can then be fused as a joint representation $\mathcal{P}^{fuse}_i$ for vertex $\mathcal{V}_i$. Notably, we achieve the $\mathcal{P}^{fuse}_i$ $\in \mathbb{R}^{{}D^{out}_f}$ through a fusion module ($\mathbb{F}$) consisting of an addition operation followed by a \textit{MLP}. Next, we adopt the K-Nearest neighbor algorithm (\textit{KNN}) on this joint representation to find the most relevant $N_n$ neighbors of $\mathcal{V}_i$ in the $\mathcal{G}$. Here, we denote the set of edges between vertex $\mathcal{V}_i$ and its neighbors as $\mathcal{E}^{nei}_i$ $\in \mathbb{R}^{N_n}$.


A direct approach to aggregating vertices’ features is to follow the methods in \cite{velickovic2018graph, 8954040}, which achieve attentive aggregation by obtaining similarity or relative matrices of features (i.e., $\mathcal{P}^{fuse}$) that define neighborhoods. However, there is one point that must be considered in our work. As shown in Figure \ref{tab: neighbor_finding}, in scenarios with complex motion states, vertices and their neighbors may be distant but semantically strongly related. We aim for the model to fully consider the motion correlation between neighbors and the central vertex when aggregating for such vertices rather than just global semantics. To this end, we enhance the contribution of coordinate attributes in graph representation for more accurate motion description by calculating aggregation weights using only coordinate clues. Further, we update the coordinate attributes with spatio-temporal relations in the neighborhood to enlarge their capability in motion description layer-by-layer. The ablations in Table Table \ref{tab:EDAL_choice} verify the efficacy of our intuitive design.


\noindent\textbf{Attentive aggregation}\indent We calculate attention scores $w.r.t$ the coordinate attribute as formulated in \ref{attn_score}.
\begin{equation}
\begin{aligned}
&{Score_i} = {\mathbb{M}^U_{A}} (\underset{j:(i,j) \in \mathcal{E}^{nei}_i}{\Gamma_{rel}} ({\mathcal{P}^U_i}, {\mathcal{P}^U_j})), 
\end{aligned}
\label{attn_score}
\end{equation}
where the function $\Gamma_{rel}$ is used for concatenating its two inputs and stacking them under the constraint ($j:(i,j) \in \mathcal{E}^{nei}_i$).
The function ${\mathbb{M}^U_{A}}$, a \textit{MLP} with the Softmax activation, is imposed for mapping the input in $\mathbb{R}^{N_n \times 2{D^{in}_f}}$ to a vector of attentive scores ${Score_i} \in \mathbb{R}^{N_n}$. Next, we can aggregate features for vertex attentively as described in \ref{feat_aggr}.
\begin{equation}
\begin{aligned}
{\mathcal{F}^{Aggr}_i} = \sum  {Score_i} * (\underset{j:(i,j) \in \mathcal{E}^{nei}_i}{\mathbb{S}}({\mathcal{P}^{fuse}_j})), 
\end{aligned}
\label{feat_aggr}
\end{equation}
where $\mathbb{S}$ works for stacking joint representations (${\mathcal{P}^{fuse}_j} \in \mathbb{R}^{D^{out}_f}$) of all vertex's neighbors and its output is in $\mathbb{R}^{N_n \times {^{out}}D_f}$. The attentive scores ${Score_i}$ can then be applied to re-weight and obtain the aggregated features ${\mathcal{F}^{Aggr}_i}$ $\in \mathcal{R}^{{}D^{out}_f}$ for vertex $\mathcal{V}_i$ through a summation operation over its neighbors. 

\noindent\textbf{Coordinate attribute update}\indent 
We depict the derivation of updating coordinate attributes by ${\mathcal{U}^{upd}_i} = {\mathbb{Q}^U_{upd}} ({Rel_i})$, where ${Rel_i}$ is obtained in \ref{attn_score}, $\mathbb{Q}^U_{upd}$ consists of an average pooling followed by a \textit{MLP} for feature aggregation and feature projection. After these two processes, the updated coordinate attribute ${\mathcal{U}^{upd}_i}$ $\in \mathbb{R}^{D^{in}_f}$ of ${\mathcal{V}_i}$ can be achieved. 
By updating the coordinate attribute of each vertex with local spatio-temporal relations with its neighbors, the vertex is equipped with spatio-temporal positions and local motion associations simultaneously, which will be transmitted to the neighbor definition and feature aggregation processes in the following layer.

\noindent\textbf{Shortcut connection}\indent \label{shortcut_connection}
In the EDAL, two shortcut connections are included for both coordinate and feature attributes of vertices in the graph. In specific, two \textit{MLP}s ($M^U_{ID}$ and $M^F_{ID}$) are applied to input attributes $\mathcal{U}_i$ and $\mathcal{F}_i$ respectively. Finally, we add the attained features from $M^U_{ID}$ and $M^F_{ID}$ to our achieved updated coordinate attribute (${\mathcal{U}^{upd}_i}$) and aggregated features (${\mathcal{F}^{Aggr}_i}$) to obtain the final output of an EDAL, $i.e.$, ${\mathcal{U}^{out}_i}$ $\in \mathbb{R}^{{D^{in}_f}}$ and ${\mathcal{F}^{out}_i}$ $\in \mathbb{R}^{{D^{out}_f}}$.

\subsubsection{Network structure}\indent
The structure of the EDGCN is the same for all datasets except for the sub-stream task head. Three EDALs are cascaded to extract discriminative features from event data sequentially. For recognition tasks, $e.g.$, object recognition and action recognition, we apply a Max pooling operation followed by a \textit{MLP} after the third EDAL and then feed the output of this \textit{MLP} to a fully connected layer for categorical prediction. As for the detection task, we follow the setting provided by \cite{Schaefer22cvpr} to apply a YOLO-based detection head \cite{redmon2016you} to our extracted event-based contexts for object detection. 

\begin{figure}[]
\centering
\includegraphics[width=1\linewidth]{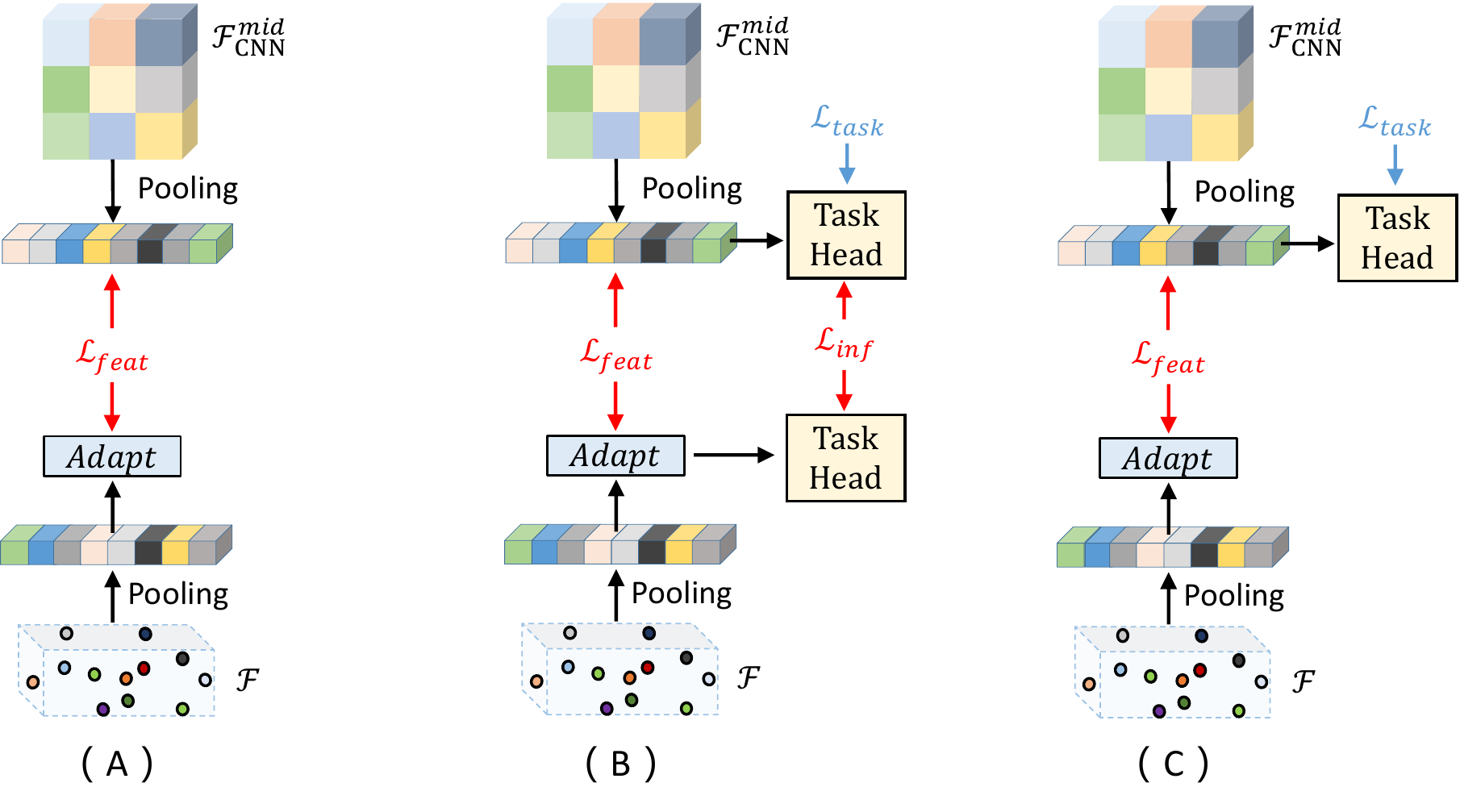}
\caption{Different choices of distillation between intermediate features from frame-based models ($\mathcal{F}^{mid}_{CNN}$) and the EDGCN ($\mathcal{F}$). The adapt module is a \textit{MLP} for dimension alignment between two feature vectors. Losses with blue color are for frame-based model training, and red color are for transfer learning. We adopt (C) as our choice and detail the reason in \ref{sec:abl}.}
\label{contrastive_choice}
\end{figure}

\subsection{Cross-representation distillation (CRD)} \label{subsec:CRLS}
As shown in \ref{pipeline}, in our cross-representation distillation framework, the teacher network is a frame-based learning branch with dense event-based representations as input and learning them starting from well-trained CNNs, while the student network is our \textit{EDGCN} model. Then, a combined distillation constraint working on different layers is tailored to fulfill this transfer.

\noindent\textbf{Distillation framework}\indent
The key issue in achieving this challenging frame-to-graph transfer is the design of a distillation structure that carefully considers the varying cross-representation discrepancy across different layers. To this end, we propose a hybrid distillation approach that includes two views. (1) Inference-level distillation of contexts from final prediction outputs. For instance, we apply the distillation loss proposed in \cite{hinton2015distilling} for classification tasks and the L1 loss \cite{Romero15-iclr} for regression-based tasks like position estimating in object detection. (2) Feature-level constraints that transfer hints from intermediate features of frame-based models. The knowledge transfer of intermediate features has been verified to effectively improve the training effect of the model to be transferred. However, for traditional CNNs and our EDGCN, their learning logic for event data from shallow to deep might be totally different. Thus, hard constraints (e.g., L1/L2 distance) \cite{pix2pix2017} would be too strict for our transfer task. For this reason, we exploit the contrastive loss, NT-Xent \cite{chen2020simple, huang2021spatio}, to realize the transfer by increasing the correlation between intermediate features from both networks. In specific, three variants to equip NT-Xent loss in the CRD are proposed in \ref{contrastive_choice}, with method \ref{contrastive_choice}.(C) being adopted as the final choice. In \ref{sec:abl}, we detail the reason for our choice and its advantages in conjunction with experimental results.

\noindent\textbf{Optimization}\indent
The whole training process of our study can be divided into two parts. First, the training process of frame-based models with task-specific loss solely is described by ${\mathcal{L}^{frame}_{total}} = \sum_{i}^{N_t} \mathcal{L}^i_{task}$, 
where $\mathcal{L}^i_{task}$ are task-specific losses. They are applied to multiple prediction layers individually (\ref{pipeline}). $N_t$ represents the number of intermediate features used for cross-representation learning. Next, we optimize our proposed EDGCN with the loss for task-specific supervision ($\mathcal{L}_{task}$), the loss for inference-level knowledge transferring ($\mathcal{L}_{inf}$) and a series of contrastive losses ($\mathcal{L}^i_{feat}$) for feature-level distillation as described in \ref{stu_training}. 
\vspace{-1.2em}
\begin{equation}
\begin{aligned}
{\mathcal{L}^{Edgcn}_{total}} = \lambda\mathcal{L}_{task} + (1-\lambda)\mathcal{L}_{inf} + \sum_{i}^{N_t} \mathcal{L}^i_{feat}, 
\end{aligned}
\label{stu_training}
\vspace{-1em}
\end{equation}
where $\lambda$ control the contribution of the first two components in the training process and set it as 0.5 for all experiments.

\section{Experimental Results}\label{sec:experi}
We evaluate our proposed method on several tasks, $e.g.$, object classification, action recognition and object detection. Besides, we validate the superiority of the EDGCN on the model complexity (measured in the number of trainable parameters) and the number of floating-point operations (FLOPs). Then, the efficacy of the proposed CRD and its generalizability on multiple vision tasks is validated.

\begin{table}[] \small
\begin{center}
\renewcommand\arraystretch{1.3}
\caption{Statistics of different datasets and parameter settings for representation construction. \footnotesize{\textit{Descriptions of datasets are placed in supplementary.}}}
\label{tab:datasets}
\begin{threeparttable}[h]
\setlength{\tabcolsep}{1.7mm}{
\begin{tabular}{ccccc}
\hline\hline
Datasets    & N-Cal \cite{orchard2015converting}    & CIF10 \cite{li2017cifar10}    & N-C \cite{sironi2018hats}  & DVS128 \cite{amir2017low} \\ \hline
Duration    & 300 ms   & 1280 ms   & 100 ms & 6000 ms\\
Classes     & 101      & 10          & 2     & 11    \\
Samples     & 8246     & 10000      & 24029 & 1342   \\
($v_x,v_y$)    & (10, 10) & (10, 10) & (5, 5)  & (5, 5) \\
$v_t$        & 25 ms    & 60 ms      & 25 ms & 40 ms  \\
$N_v$     & 2048     & 2048     & 512     & 512    \\\hline\hline  
\end{tabular}}
\end{threeparttable}
\end{center}
\vspace{-2em}
\end{table}

\noindent\textbf{Implementation details}\indent
We choose ResNet \cite{he2016deep} as the backbone of frame-based models used for applying CRD to object classification and object detection tasks and adopt I3D-R (w/ ResNet50) \cite{chen2021deep} for the action recognition task, where ResNet is pretrained on ImageNet and I3D-R is pretrained using Kinetics-400 dataset. We train them using the Adam optimizer with batch size 32 and an initial learning rate (\textit{lr}) of 1e-4, which is reduced by a factor of 2 after 20 epochs. The dense input of these frame-based models is VoxelGrid \cite{zihao2018unsupervised}. For the EDGCN, we keep its network structure (\ref{pipeline}) unchanged for all datasets except for its task head. We use SGD optimizer with an initial \textit{lr} of 1e-1 for object classification and action recognition, and reduce the \textit{lr} until 1e-4 using cosine annealing. We choose Adam optimizer with batch size 32 for detection, and reduce \textit{lr} starting from 1e-2 by a factor of 2 after 20 epochs. The settings are consistent for training the EDGCN solely and with CRD. We list the statistics of adopted datasets and their settings of voxel-wise representation construction. We average over five runs as our final results for all experiments.
\textit{We release the code and network settings in the supplementary.}

\subsection{Object classification}
Event-based object classification is an essential application since event cameras can recognize objects more accurately than traditional cameras in scenarios with severe motion blur and extreme lighting conditions. In this work, we select three challenging datasets commonly used for evaluating event-based object classification, $i.e.$, N-Cal \cite{orchard2015converting}, N-C \cite{sironi2018hats}, and CIF10 \cite{li2017cifar10} (Table Table \ref{tab:datasets}). The ResNet-18 is the backbone of the frame-based model that we employed for optimizing EDGCN with CRD, and its performance on N-Cal, N-C, and CIF10 is $0.868$, $0.964$, and $0.757$. The cross-entropy loss is utilized as $\mathcal{L}_{task}$ for the model's training.

\begin{table}[]\small
\begin{center}
\renewcommand\arraystretch{1.3}
\caption{Comparison of models $w.r.t$ classification accuracy. \footnotesize{\textit{${\dagger}$: Performance of the EDGCN trained solely. ${\ddagger}$: Performance of the model trained with CRD. ${\S}:$ F:frame-based method; P:point-based method.}}}
\label{tab:classification}
\begin{threeparttable}[h]
\setlength{\tabcolsep}{2.5mm}{
\begin{tabular}{lcccc}
\hline\hline
Method & Type$^{\S}$  & N-Cal & N-C & CIF10 \\ \hline
\multicolumn{5}{c}{Pretrained on ImageNet \cite{deng2009imagenet}} \\ \hline
EST \cite{gehrig2019end}       & F     & 0.837        & 0.925   & 0.749       \\
M-LSTM \cite{Cannici_2020_ECCV}      & F        & {0.857}               & {0.957}           &{0.730}     \\
MVF-Net \cite{mvfnet}       & F   & \textbf{0.871} & \textbf{0.968} & \textbf{0.762}  \\ \hline
\multicolumn{5}{c}{Without pretraining} \\ \hline
EST \cite{gehrig2019end}           & F       & {0.753}           & 0.919  & 0.634        \\
M-LSTM \cite{Cannici_2020_ECCV}   & F  & 0.738               & 0.927          & 0.631     \\
MVF-Net \cite{mvfnet}     & F    & 0.687 & 0.927  & 0.599 \\
AsyNet \cite{messikommer2020event}   & F  & 0.745 & {0.944}  & {0.663}  \\
EventNet \cite{Sekikawa_2019_CVPR}    & P & 0.425  & 0.750 & 0.171  \\
RG-CNNs \cite{Graph-based}            & P & 0.657 & 0.914 & 0.540 \\ 
EvS-S \cite{Li21iccv}            & P & 0.761 & 0.931 & 0.680 \\ 
EV-VGCNN \cite{Deng_2022_CVPR}            & P & 0.748 & 0.953 & 0.670 \\ 
AEGNN \cite{Schaefer22cvpr}             & P & 0.668 & 0.945 & - \\ 
VMV-GCN \cite{Xie_2022_RAL}             & P & 0.778 & 0.932 & 0.690 \\ 
EV-Transformer \cite{li2022event}             & P & 0.789 & 0.954 & 0.709 \\ \hline
\textbf{Ours}$^{\dagger}$     & P   & \textbf{0.801}      & \textbf{0.958}  & \textbf{0.716}      \\ 
\textbf{Ours w/ CRD}$^{\ddagger}$    & P   & \textbf{0.835}      & \textbf{0.963}     & \textbf{0.752}      \\ \hline\hline
\end{tabular}}
\end{threeparttable}
\end{center}
\vspace{-2em}
\end{table}

\noindent\textbf{Classification accuracy}\indent
We compare the proposed method with SOTA methods falling in both point-based and frame-based categories. The Table \ref{tab:classification} presents that methods with graph-based learning (RG-CNNs, Evs-S, EV-VGCNN, AEGNN, VMV-GCN, EV-Transformer and ours) are prevalent $w.r.t$ classification accuracy over other point-based methods. Notably, the proposed EDGCN achieves top performance among these graph-based approaches, revealing the effectiveness of our proposed learning model. We attribute these improvements to the EDAL that can aggregate features for vertices considering all attributes dynamically, which allows us to efficiently and precisely extract the semantics of events. 

Excitedly, the introduced CRD can improve the performance of the EDGCN by a large margin. It proves that our CRD can successfully utilize the pretrained weights of CNNs to ease the learning of our graph model and improve its representation ability, even with large image-to-graphs gaps. However, our model, largely improved by the CRD scheme, still lags behind some frame-based methods ($e.g.$, the MVF-Net) that are with pretraining. We attribute this to the much smaller discrepancy between image-frame than our frame-graph, allowing those frame-based methods to directly use the model weights well well-trained with large-scale datasets.
\par Moreover, Table \ref{tab:model_complexity} shows that our method holds large advantages in model complexity and computational cost over other approaches. 

\subsection{Action recognition}

In this section, we choose the action recognition task to validate the advantages of our model in encoding motions using the DVS128 \cite{amir2017low} dataset which contains samples derived by different gestures. We follow \cite{Graph-based} to sample all test data with 0.5s duration. The cross-entropy loss is utilized as $\mathcal{L}_{task}$ for the model's training. The I3D-R is chosen as the backbone of the frame-based teacher network when performing CRD for optimizing EDGCN and the performance of the teacher on DVS128 is $0.981$.

\begin{table}[]\small
\begin{center}
\renewcommand\arraystretch{1.3}
\caption{Comparison of models on the DVS128 dateset.}
\begin{threeparttable}[h]
\setlength{\tabcolsep}{1.4mm}{
\begin{tabular}{lcccc}
\hline\hline
Method & Type & Accuracy & GFLOPs & \#Params \\ \hline
LIAF-Net \cite{LIAF-Net} & F & 0.976 & 13.6 & - \\
TA-SNN \cite{yao2021temporal} & F & \textbf{0.986} & - & - \\ \hline
RG-CNN (Res.3D) \cite{Graph-based} & P & 0.972 & 13.72 & 12.43 M \\ 
EV-VGCNN \cite{Deng_2022_CVPR}      & P      & 0.959 & 0.46 & 0.82 M \\
VMV-GCN \cite{Xie_2022_RAL}       & P      & 0.975 & 0.33 & 0.84 M \\ \hline
\textbf{Ours}    & P             & 0.985         & \textbf{0.14}   & \textbf{0.72 M}       \\
\textbf{Ours w/ CRD}   & P      & 0.983 & \textbf{0.14}   & \textbf{0.72 M}       \\ 
\hline\hline
\end{tabular}}
\label{tab:action}
\end{threeparttable}
\end{center}
\vspace{-2em}
\end{table}

\noindent\textbf{Recognition performance}\indent
Table \ref{tab:action} shows that our model has significant advantages over other point-based approaches $w.r.t$ recognition accuracy, model complexity, and computational cost. Despite using highly sparse input data, our model achieves accuracy comparable to the state-of-the-art frame-based method (TA-SNN). Notably, I3D-R exhibits weakness in this task compared to EDGCN even with pretraining and a much heavier model. 

These findings suggest that EDGCN can accurately extract motion cues while preserving the sparsity of event data. This advantage can be attributed to two views. ($i$) Accurate neighborhood definition. ($ii$) Attentive aggregation which relies only on spatio-temporal relations for augmenting motion elements in feature representation. We also evaluated our work with aggregation considering joint features ($\mathcal{P}^{fuse}$) instead of only coordinates ($\mathcal{P}^{\mathcal{U}}$). The results on multiple tasks are consistent with our expectations (Table \ref{tab:EDAL_choice}).




\subsection{Object detection}

Event-based object detection is an emerging topic to simultaneously solve object localization and categorization. This task requires event-based models with powerful semantics and motion encoding capabilities. We conduct experimental comparisons for this task on the N-Cal dataset, which is a single object detection dataset containing 101 classes. The ResNet-34 is used as the backbone of the frame-based teacher branch for our CRD, and its performance on N-Cal is $0.76$. Following the setup in \cite{Schaefer22cvpr}, we use a collection of losses (a weighted sum of class, bounding box offset and shape as well as prediction confidence losses) as the $\mathcal{L}_{task}$ for training.

\begin{table}[]\small
\begin{center}
\renewcommand\arraystretch{1.3}
\caption{\small Comparison of models $w.r.t$ the eleven-point mean average precision (mAP) on the event-based object detection task.}
\setlength{\tabcolsep}{2.5mm}{
\begin{tabular}{lccc}
\hline\hline
Methods   & Type  & N-Cal (mAP$\uparrow$) \\ \hline 
YOLE  \cite{cannici2019asynchronous}    & F    &  0.398  \\  
Asynet \cite{messikommer2020event}   & F    &   0.643  \\  
NvS-S  \cite{Li21iccv}   & P   &  0.346    \\  
AEGNN  \cite{Schaefer22cvpr}   & P   &  0.595     \\ \hline 
\textbf{Ours}      & P   &  \textbf{0.657}     &      \\ 
\textbf{Ours w/ CRD} & P   &  \textbf{0.711}    &      \\ 
\hline\hline
\end{tabular}}
\label{tab:detection}
\end{center}
\vspace{-1.5em}
\end{table}

\noindent\textbf{Detection performance}\indent
We utilize the eleven-point mean average precision (mAP) to measure our models on the detection task. From results in Table \ref{tab:detection} and Table \ref{tab:model_complexity}, we conclude that our approach achieves large improvement on mAP over others with much fewer parameters and computational costs. More importantly, the proposed EDGCN trained solely exceeds other graph-based models such as NvS-S and AEGNN by a large margin, indicating the superiority of our EDGCN. Besides, CRD also largely boosts the detection performance in addition to object and action recognition tasks, suggesting the well generalizability of our proposed learning scheme.

\subsection{Complexity analysis}

\begin{table}[]\small
\begin{center}
\renewcommand\arraystretch{1.3}
\caption{\small Comparison of models on the model complexity (\#Params) and the number of FLOPs.}
\setlength{\tabcolsep}{1.5mm}{
\begin{tabular}{lcccc}
\hline\hline
Method  & Type & \#Params  & GFLOPs  & Time\\ \hline
EST \cite{gehrig2019end}             & F      & 21.38 M    &  4.28    & 6.41 ms     \\
M-LSTM \cite{Cannici_2020_ECCV}      & F  & 21.43 M    &  4.82     & 10.89 ms \\
MVF-Net \cite{mvfnet}                & F & 33.62 M    & 5.62    & 10.09 ms     \\ 
AsyNet \cite{messikommer2020event}   & F           & 3.69 M & 0.88  & - \\ \hline
EventNet \cite{Sekikawa_2019_CVPR}   & P        & 2.81 M    & 0.91    & \textbf{3.35 ms}      \\ 
PointNet++ \cite{wang_2018_wacv}     & P    & 1.76 M    &   4.03   & 103.85 ms  \\ 
RG-CNNs \cite{Graph-based}           & P  & 19.46 M    &  0.79     & -  \\
EV-VGCNN \cite{Deng_2022_CVPR}       & P                &  0.84 M    &  0.70   & 7.12 ms      \\ 
AEGNN \cite{Schaefer22cvpr}          & P              & 20.4 M    & 0.75  & -      \\ 
VMV-GCN \cite{Xie_2022_RAL}          & P              &  0.86 M    &  1.30   &  6.27 ms      \\ \hline 
\textbf{Ours}                        & P   &  \textbf{ 0.77 M}    &  \textbf{0.57}   & 3.84 ms      \\ \hline\hline
\end{tabular}}
\label{tab:model_complexity}
\end{center}
\vspace{-2em}
\end{table}

We compute the complexity and FLOPs of object classification models on the N-Cal dataset following \cite{Graph-based, Deng_2022_CVPR}. Results in Table \ref{tab:model_complexity} show that our approach holds extremely low model complexity and computational cost, indicating the high efficiency the learning system holds in extracting representative features from event data. We also measure inference time on the N-C dataset using PyTorch on an Nvidia RTX 3090. Our approach achieves leading performance with only 3.84 ms processing time per sample (equivalent to 260 Hz frame-rate), demonstrating its practical value in high-speed scenarios. Although slightly slower than EventNet, our method’s higher performance demonstrates its better efficacy and generalization ability.


\subsection{Ablation study} \label{sec:abl}
\begin{table}[]\small
\begin{center}
\renewcommand\arraystretch{1.3}
\caption{\small The ablation study on the effects of different designs to model's performance. \footnotesize{\textit{AA$_{U}$: attentive aggregation using coordinate attributes. AA$_{fuse}$: attentive aggregation using the fused joint representation. UPD: coordinate attribute update module.}}}
\begin{threeparttable}[h]
\setlength{\tabcolsep}{0.7mm}{
\begin{tabular}{c|c|c|c|c|c|c|cccc}
\hline\hline
& AA$_{U}$ & UPD & AA$_{fuse}$ & $\mathcal{P}^U$ & $\mathcal{P}^F$ & $\mathcal{P}^{fuse}$ & N-Cal & CIF10 & DVS128\\ \hline
A & $\checkmark$ & $\checkmark$ &                                & $\checkmark$ &              &              &  0.793            &  0.695    &  0.954     \\
B & $\checkmark$ & $\checkmark$ &                                 &              & $\checkmark$ &              &  0.790            &  0.701     &  0.970     \\ \hline
C & $\checkmark$ & $\checkmark$ &                                &              &              & $\checkmark$ & \textbf{0.801}   & \textbf{0.716}      & \textbf{0.985}       \\ \hline
D & $\checkmark$ &  &                               &              &              & $\checkmark$ &   0.783           & 0.694      &  0.972     \\ 
E &  & $\checkmark$ & $\checkmark$                               &              &              & $\checkmark$ &   0.788           &  0.703     & 0.969     \\ \hline\hline
\end{tabular}
}
\label{tab:EDAL_choice}
\end{threeparttable}
\end{center}
\end{table}

\begin{table}[]\small
\renewcommand\arraystretch{1.3}
\caption{\small The ablation study on the effects of different designs to model's performance.
}
\begin{threeparttable}[h]
\setlength{\tabcolsep}{2.5mm}{
\begin{tabular}{cccccccc}
\hline\hline
EDGCN             &        $\mathcal{L}_{inf}$                                     & \multicolumn{4}{c}{$\mathcal{L}_{feat}$}                  & N-Cal      & CIF10      \\ \cline{3-6}
        &  & A            & B            & C            & D       &  &  \\ \hline
$\checkmark$ &                                            &             &              &              &              & 0.801     & 0.716      \\
$\checkmark$ & $\checkmark$                               &              &              &              &              &  0.818     &   0.724    \\
$\checkmark$ & $\checkmark$                               & $\checkmark$ &              &              &              &  0.825     &   0.733    \\
$\checkmark$ & $\checkmark$                               &              & $\checkmark$ &              &              &  0.830     &   0.749    \\
$\checkmark$ & $\checkmark$                               &              &              & $\checkmark$ &              &  \textbf{0.835}     & \textbf{0.752}      \\
$\checkmark$ & $\checkmark$                               &              &              &              & $\checkmark$ &  0.814     &  0.721     \\ \hline\hline
\end{tabular}
}
\label{tab:transfer_choice}
\end{threeparttable}
\vspace{-0.5em}
\end{table}

In this section, we evaluate our method through various settings, where setup in Table \ref{tab:EDAL_choice} is for evaluating core modules of EDAL and Table \ref{tab:transfer_choice} is for verify the the contribution of each constraint in the proposed CRD. A visualization is also presented to show the benefits of CRD in improving the representation ability of EDGCN. \textit{Please refer to the supplementary for more ablation studies and discussions}. 


\noindent\textbf{Effectiveness of learning modules in EDAL}\indent We investigate the effectiveness of our neighborhood definition method through settings A, B, and C in Table \ref{tab:EDAL_choice}, where A, B, and C represent defining neighbors using only coordinate attributes ($\mathcal{P}^U$), only semantic features ($\mathcal{P}^F$), and a joint representation ($\mathcal{P}^{fuse}$) of all attributes respectively. The results show that our neighborhood definition method consistently improves model performance across different datasets. In particular, compared to the method of finding neighbors using only coordinates as done in \cite{Graph-based, Deng_2022_CVPR, Li21iccv, Schaefer22cvpr, Xie_2022_RAL, 9181247}, our method achieves significant improvement in action recognition tasks. This confirms our observation in Figure \ref{tab: neighbor_finding} that it is difficult to find neighbors highly related to a vertex’s motion and semantic information based solely on coordinates in scenarios where motion states are complex. Additionally, we explore the impact of attentive aggregation mode on model performance (C and E). Experiments show that by enhancing the contribution of motion information during aggregation, our method can obtain more efficient graph representation. This validates the rationality of our design. Also, we can see that UPD further brings considerable improvement, indicating that the coordinate attribute strengthened by UPD can further facilitate the aggregation process.

\noindent\textbf{Designs in the CRD}\indent 
In \ref{contrastive_choice}, we introduce three intermediate feature-level knowledge transfer designs, A, B, and C, which all employ the contrastive loss to ensure consistency between teacher and student features. Variant B, however, differs from A by adding task-specific loss to both student and teacher branches. In contrast, C only applies the task-specific loss to the teacher branch. Moreover, D in Table \ref{tab:transfer_choice} replaces the contrastive loss in A with hard constraints such as L1 loss \cite{Romero15-iclr} for feature distillation.


Table \ref{tab:transfer_choice} shows that inference-level distillation outperforms the EDGCN trained solely. Furthermore, A, B, and C variants boost EDGCN by supervising intermediate features, with B being the best. This is due to $\mathcal{L}_{task}$ on intermediate features, which makes the teacher represent events more comprehensively and provide better guidance to the student. Surprisingly, B is worse than C with extra task loss for the student, potentially due to the limited power of intermediate features for reliable prediction. Furthermore, variant D degrades the entire learning framework, likely because of different learning logic of two models. This supports our choice of contrastive loss for frame-to-graph feature distillation, as different feature dimensions may have different semantics and hard constraints are not good for feature consistency \cite{Islam_2021_ICCV}.

\begin{figure}[]
\centering
\includegraphics[width=1\linewidth]{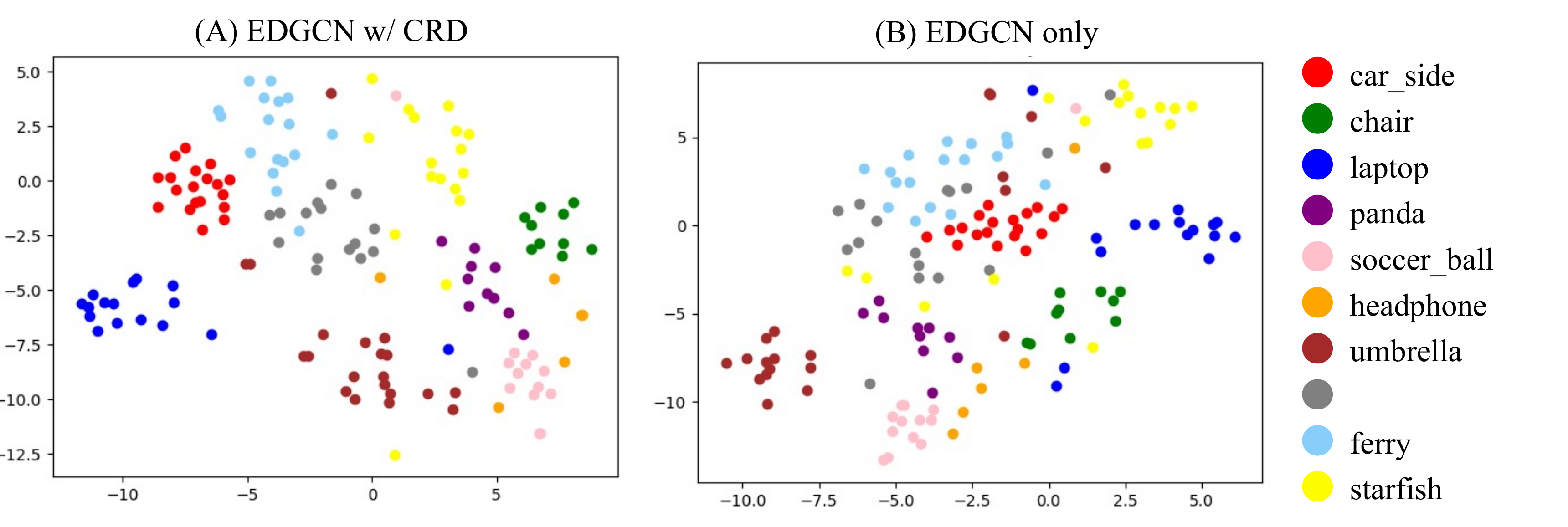}
\caption{The t-SNE visualization on the test set of N-Cal. 
}
\label{tsne}
\end{figure}

\noindent\textbf{Visualization for feature representation}\indent We further illustrate t-SNE of features on the N-Cal dataset in \ref{tsne}. The EDGCN trained solely shows weakness in achieving explicit decision boundaries among some challenging classes such as car\_side and ferry due to their similar appearances (\ref{tsne}.(B)). This limitation of EDGCN can be mitigated by CRD, where more discriminative features are obtained (\ref{tsne}.(A)), indicating the efficacy of CRD in boosting the representation ability of EDGCN. 

\section{Conclusion}\label{sec:conclu}
We propose a novel event-based GCN (EDGCN) for event-based learning that precisely defines the neighborhood for each event-based vertex considering all its attributes and dynamically updates vertex attributes layer-by-layer. We also introduce a cross-representation distillation framework (CRD) for point-based methods that leverages large-scale prior from frame-based models to facilitate EDGCN training. Comprehensive experiments on various vision tasks validate the efficacy of our EDGCN and CRD. Since CRD has potential for migrating to other point-based methods for event data, we argue that this learning strategy may open new research avenues for event-based model learning.

\appendices

\section{Network structure}\label{sec:network}

\begin{table}[t]
\renewcommand\arraystretch{1.3}
\caption{Settings of network structures.}
\centering
\setlength{\tabcolsep}{3mm}{
\begin{tabular}{c|c|c|c}
\hline\hline
 & Layer 1 & Layer 2 & Layer 3  \\ \hline
${^{in}D_u}$               &   3        &     25      &    64    \\
${^{in}D_f}$               &   25        &    64       &   64      \\
${^{out}}D_f$              &   64        &     64      &   128  \\
${N_n}$                    &   20      &   20      &  20  \\\hline\hline
\end{tabular}}
\label{network_struct}
\end{table}

As stated in the main manuscript, we keep the network structure of the proposed EDGCN unchanged for all datasets except for its task head. Here, we detail the structure settings of three feature aggregation layers (EDALs) in our model (Eq. \ref{network_struct}). 

\section{Robustness to the time span of input events} \label{sec:robust_latency}
\begin{figure}[h]
\begin{center}
\includegraphics[width=1\linewidth]{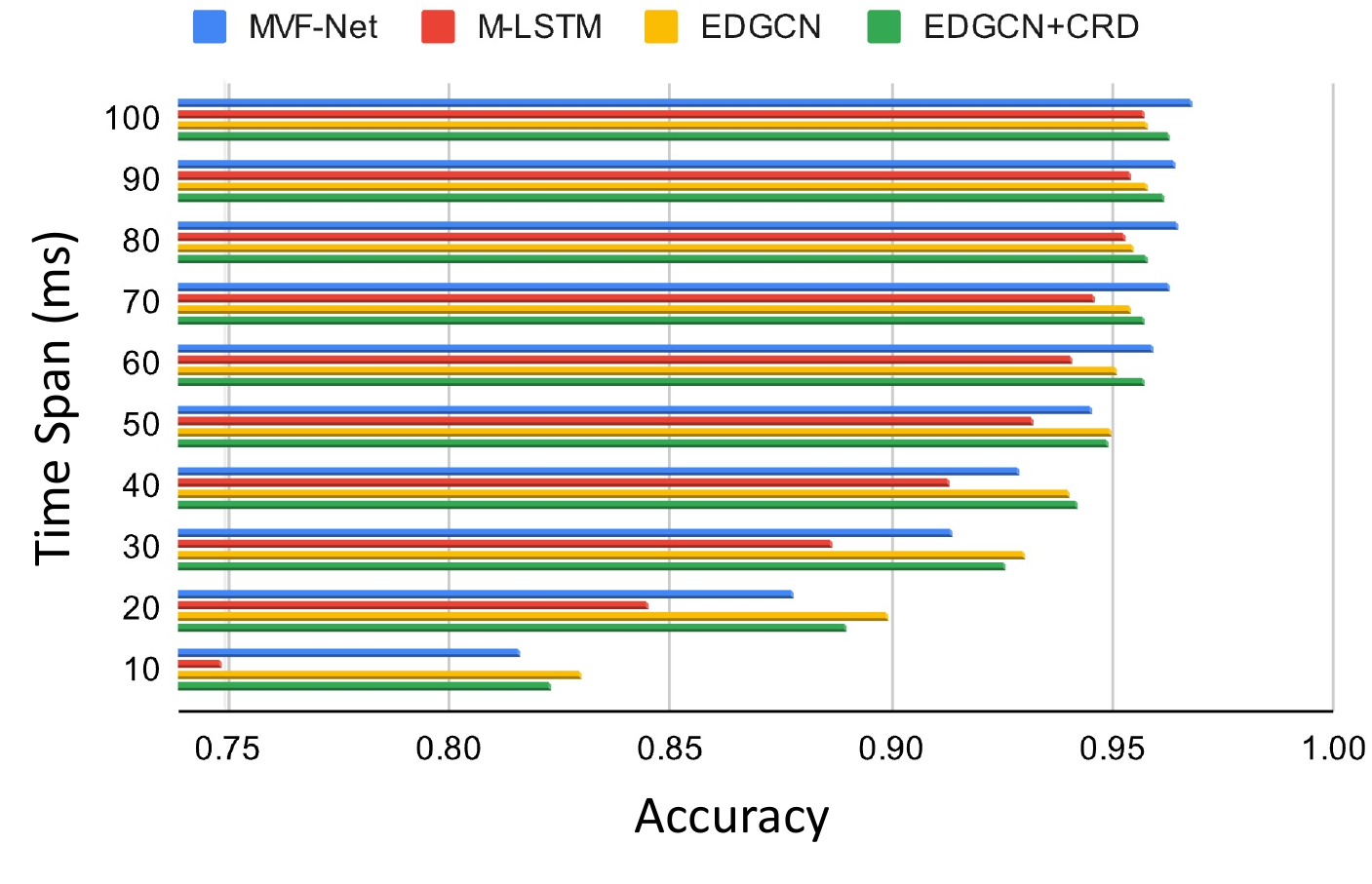}
\end{center}
  \caption{Analysis of event-based methods $w.r.t$ the time span of input events and classification accuracy. Best viewed in color.}
\label{latency}
\end{figure}

We follow the experiment settings adopted in \cite{sironi2018hats, Cannici_2020_ECCV} to validate the robustness of our approach to the time span of input events. For instance, if a learning model takes event signals with 100ms length as input, the time span of input events should be 100ms. We compare our model with two SOTA frame-based approaches (MVF-Net \cite{mvfnet}, and M-LSTM \cite{Cannici_2020_ECCV}) on a series of testing settings using the N-C dataset and present results in Fig. ref{latency}. The testing set contains event samples with variable time duration ($10\% \sim 100\%$). We can see that though the best performance of our model is slightly lower than frame-based approaches, the robustness of our model to the time span of input events is better than their methods, especially with the short time span input, showing the reliability of our method in processing event data under complex working conditions.

\begin{figure*}[h]
\begin{center}
\includegraphics[width=1\linewidth]{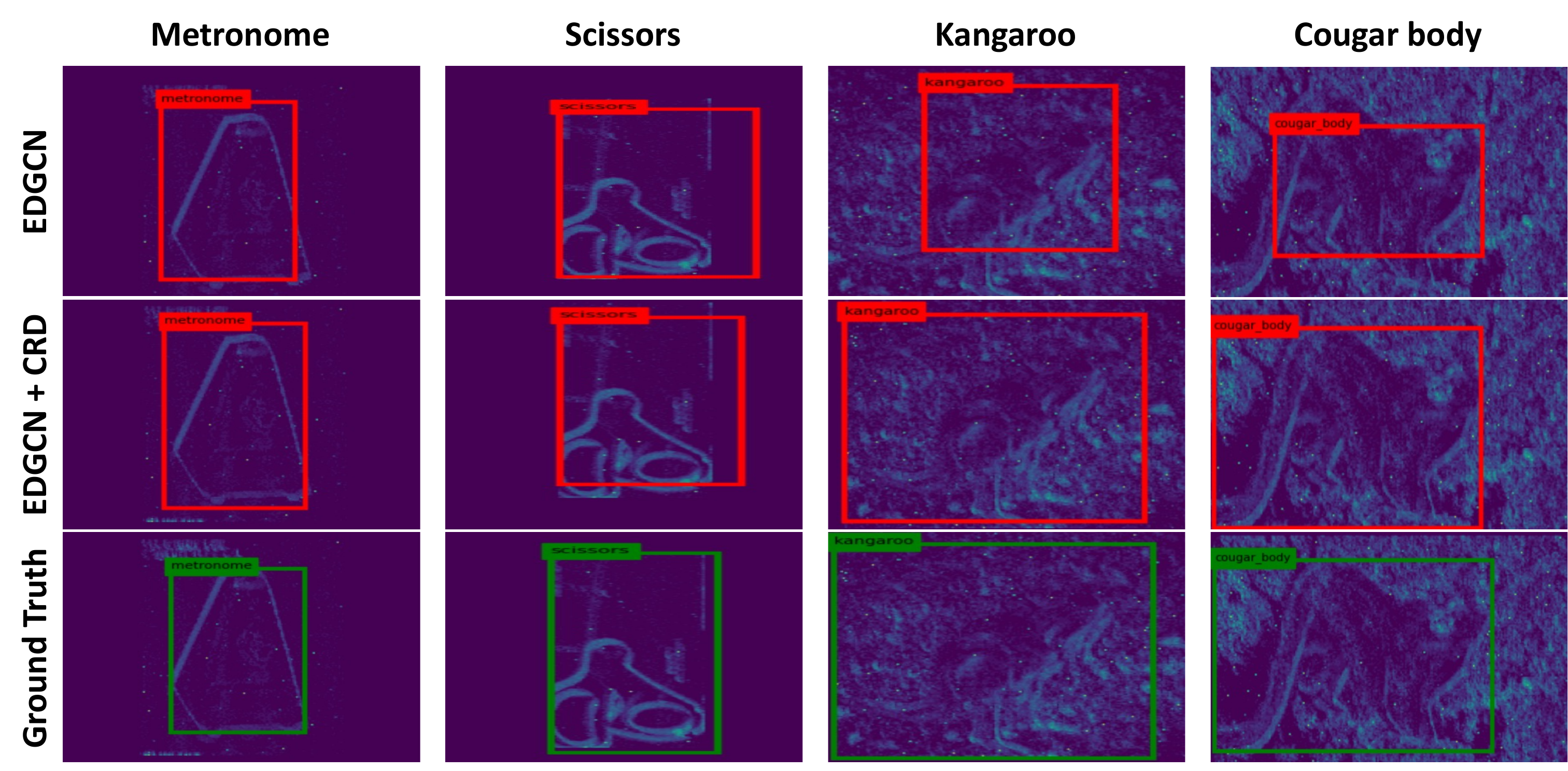}
\end{center}
  \caption{Qualitative results of the detection task performed by the proposed EDGCN and CRD on N-Cal \cite{orchard2014bioinspired} dataset. Specifically, we show examples correctly categorized by the model and analyze their object localization ability. It can be seen that, although it is easy for EDGCN trained solely to recognize and localize the simple objects in the first two columns, there exists hardness to do the same thing on the event-based samples with complex background textures. In the last two columns, we show that after applying the CRD, we have mitigated to a certain extent the adverse effects on network localization caused by the interference of complex background textures, illustrating the promotion brought by the CRD on feature learning of the EDGCN. Best viewed in color.}
\label{detection}
\end{figure*}

\section{Compare to most relative studies.}\label{sec:discuss}
The most relative works to our approach are VGCNN \cite{Deng_2022_CVPR} and VMV-GCN \cite{Xie_2022_RAL} since their studies adopt the same type of input representations as our approach. However, unlike their works with biased feature aggregation (e.g., merely considering spatiotemporal distances in \cite{Xie_2022_RAL}) or handcrafted scaling weights (e.g., \cite{Deng_2022_CVPR}), we remove this bottleneck by considering all attributes in a learnable manner. In specific, \cite{Deng_2022_CVPR} uses original input coordinates to find and aggregate neighboring vertices while original coordinates may not be able to distinguish the proximity of vertices in the feature space. To solve the shortcoming of \cite{Deng_2022_CVPR}, \cite{Xie_2022_RAL} defines neighborhood space by updating coordinate attributes. Yet, \cite{Xie_2022_RAL} only relies on the geometric relationship between vertices and ignores the semantic relationship. The major novelty of EDGCN is how to define neighborhood space for each vertex by considering all attributes based on semantics \& spatial-temporal features adaptively. Besides, the proposed EDGCN introduces attentive feature aggregation for neighbor feature embedding, which is more flexible than \cite{Xie_2022_RAL} which just uses a Max pooling for hard aggregating. In addition, we propose a novel distillation scheme for point-based approaches which is capable of improving the performance of our approach step further.


%









\bibliographystyle{IEEEtran}
\bibliography{IEEEabrv,egbib}

\begin{thebibliography}{10}
\providecommand{\url}[1]{#1}
\csname url@samestyle\endcsname
\providecommand{\newblock}{\relax}
\providecommand{\bibinfo}[2]{#2}
\providecommand{\BIBentrySTDinterwordspacing}{\spaceskip=0pt\relax}
\providecommand{\BIBentryALTinterwordstretchfactor}{4}
\providecommand{\BIBentryALTinterwordspacing}{\spaceskip=\fontdimen2\font plus
\BIBentryALTinterwordstretchfactor\fontdimen3\font minus
  \fontdimen4\font\relax}
\providecommand{\BIBforeignlanguage}[2]{{%
\expandafter\ifx\csname l@#1\endcsname\relax
\typeout{** WARNING: IEEEtran.bst: No hyphenation pattern has been}%
\typeout{** loaded for the language `#1'. Using the pattern for}%
\typeout{** the default language instead.}%
\else
\language=\csname l@#1\endcsname
\fi
#2}}
\providecommand{\BIBdecl}{\relax}
\BIBdecl

\bibitem{gallego2019event}
G.~{Gallego}, T.~{Delbruck}, G.~M. {Orchard}, C.~{Bartolozzi}, B.~{Taba},
  A.~{Censi}, S.~{Leutenegger}, A.~{Davison}, J.~{Conradt}, K.~{Daniilidis},
  and D.~{Scaramuzza}, ``Event-based vision: A survey,'' \emph{IEEE Trans.
  Pattern Anal. Mach. Intell.}, pp. 1--1, 2020.

\bibitem{gehrig2019end}
D.~Gehrig, A.~Loquercio, K.~G. Derpanis, and D.~Scaramuzza, ``End-to-end
  learning of representations for asynchronous event-based data,'' in
  \emph{IEEE/CVF Int. Conf. Comput. Vis.}, 2019, pp. 5633--5643.

\bibitem{Cannici_2020_ECCV}
M.~Cannici, M.~Ciccone, A.~Romanoni, and M.~Matteucci, ``A differentiable
  recurrent surface for asynchronous event-based data,'' in \emph{Eur. Conf.
  Comput. Vis.}, August 2020.

\bibitem{messikommer2020event}
N.~Messikommer, D.~Gehrig, A.~Loquercio, and D.~Scaramuzza, ``Event-based
  asynchronous sparse convolutional networks,'' in \emph{Eur. Conf. Comput.
  Vis.}\hskip 1em plus 0.5em minus 0.4em\relax Springer, 2020, pp. 415--431.

\bibitem{mvfnet}
Y.~Deng, H.~Chen, and Y.~Li, ``Mvf-net: A multi-view fusion network for
  event-based object classification,'' \emph{IEEE Trans. Circuits Syst. Video
  Technol.}, pp. 1--1, 2021.

\bibitem{Tore2022pami}
R.~Baldwin, R.~Liu, M.~M. Almatrafi, V.~K. Asari, and K.~Hirakawa,
  ``Time-ordered recent event (tore) volumes for event cameras,'' \emph{IEEE
  Trans. Pattern Anal. Mach. Intell.}, pp. 1--1, 2022.

\bibitem{ev2vid}
H.~Rebecq, R.~Ranftl, V.~Koltun, and D.~Scaramuzza, ``High speed and high
  dynamic range video with an event camera,'' \emph{IEEE Trans. Pattern Anal.
  Mach. Intell.}, pp. 1--1, 2019.

\bibitem{qi2017pointnet}
C.~R. Qi, H.~Su, K.~Mo, and L.~J. Guibas, ``Pointnet: Deep learning on point
  sets for 3d classification and segmentation,'' in \emph{IEEE Conf. Comput.
  Vis. Pattern Recog.}, 2017, pp. 652--660.

\bibitem{wang_2018_wacv}
Q.~Wang, Y.~Zhang, J.~Yuan, and Y.~Lu, ``Space-time event clouds for gesture
  recognition: From rgb cameras to event cameras,'' in \emph{IEEE Winter Conf.
  on Appl. of Comput. Vis.}, 2019, pp. 1826--1835.

\bibitem{Sekikawa_2019_CVPR}
Y.~Sekikawa, K.~Hara, and H.~Saito, ``Eventnet: Asynchronous recursive event
  processing,'' in \emph{IEEE Conf. Comput. Vis. Pattern Recog.}, June 2019.

\bibitem{Graph-based}
Y.~{Bi}, A.~{Chadha}, A.~{Abbas}, E.~{Bourtsoulatze}, and Y.~{Andreopoulos},
  ``Graph-based spatio-temporal feature learning for neuromorphic vision
  sensing,'' \emph{IEEE Trans. Image Process.}, pp. 1--1, 2020.

\bibitem{Li21iccv}
Y.~Li, H.~Zhou, B.~Yang, Y.~Zhang, Z.~Cui, H.~Bao, and G.~Zhang, ``Graph-based
  asynchronous event processing for rapid object recognition,'' in
  \emph{IEEE/CVF Int. Conf. Comput. Vis.}, 2021, pp. 914--923.

\bibitem{Schaefer22cvpr}
S.~Schaefer, D.~Gehrig, and D.~Scaramuzza, ``Aegnn: Asynchronous event-based
  graph neural networks,'' in \emph{IEEE Conf. Comput. Vis. Pattern Recog.},
  2022.

\bibitem{Deng_2022_CVPR}
Y.~Deng, H.~Chen, H.~Liu, and Y.~Li, ``A voxel graph cnn for object
  classification with event cameras,'' in \emph{IEEE Conf. Comput. Vis. Pattern
  Recog.}, June 2022, pp. 1172--1181.

\bibitem{Xie_2022_RAL}
B.~Xie, Y.~Deng, Z.~Shao, H.~Liu, and Y.~Li, ``Vmv-gcn: Volumetric multi-view
  based graph cnn for event stream classification,'' \emph{IEEE Robot. Autom.
  Lett.}, vol.~7, no.~2, pp. 1976--1983, 2022.

\bibitem{li2022event}
Z.~Li, M.~S. Asif, and Z.~Ma, ``Event transformer,'' \emph{arXiv preprint
  arXiv:2204.05172}, 2022.

\bibitem{9181247}
J.~Chen, J.~Meng, X.~Wang, and J.~Yuan, ``Dynamic graph cnn for event-camera
  based gesture recognition,'' in \emph{2020 IEEE International Symposium on
  Circuits and Systems (ISCAS)}, 2020, pp. 1--5.

\bibitem{chen2020simple}
T.~Chen, S.~Kornblith, M.~Norouzi, and G.~Hinton, ``A simple framework for
  contrastive learning of visual representations,'' in \emph{Int. Conf. Mach.
  Learn}.\hskip 1em plus 0.5em minus 0.4em\relax PMLR, 2020, pp. 1597--1607.

\bibitem{hinton2015distilling}
G.~Hinton, O.~Vinyals, and J.~Dean, ``Distilling the knowledge in a neural
  network,'' in \emph{NIPS Deep Learning and Representation Learning Workshop},
  2015.

\bibitem{Romero15-iclr}
A.~Romero, N.~Ballas, S.~E. Kahou, A.~Chassang, C.~Gatta, and Y.~Bengio,
  ``Fitnets: Hints for thin deep nets,'' in \emph{Int. Conf. Learn.
  Represent.}, 2015.

\bibitem{zhang2021event}
X.~Zhang, W.~Liao, L.~Yu, W.~Yang, and G.-S. Xia, ``Event-based synthetic
  aperture imaging with a hybrid network,'' in \emph{IEEE Conf. Comput. Vis.
  Pattern Recog.}, 2021, pp. 14\,235--14\,244.

\bibitem{Zhu_2022_CVPR}
L.~Zhu, X.~Wang, Y.~Chang, J.~Li, T.~Huang, and Y.~Tian, ``Event-based video
  reconstruction via potential-assisted spiking neural network,'' in \emph{IEEE
  Conf. Comput. Vis. Pattern Recog.}, June 2022, pp. 3594--3604.

\bibitem{zihao2018unsupervised}
A.~Zihao~Zhu, L.~Yuan, K.~Chaney, and K.~Daniilidis, ``Unsupervised event-based
  optical flow using motion compensation,'' in \emph{Eur. Conf. Comput. Vis.},
  2018, pp. 0--0.

\bibitem{hagenaars2021self}
J.~Hagenaars, F.~Paredes-Vall{\'e}s, and G.~De~Croon, ``Self-supervised
  learning of event-based optical flow with spiking neural networks,''
  \emph{Advances in Neural Information Processing Systems}, vol.~34, pp.
  7167--7179, 2021.

\bibitem{Hu_2022_CVPR}
L.~Hu, R.~Zhao, Z.~Ding, L.~Ma, B.~Shi, R.~Xiong, and T.~Huang, ``Optical flow
  estimation for spiking camera,'' in \emph{IEEE Conf. Comput. Vis. Pattern
  Recog.}, June 2022, pp. 17\,844--17\,853.

\bibitem{mitrokhin2020learning}
A.~Mitrokhin, Z.~Hua, C.~Fermuller, and Y.~Aloimonos, ``Learning visual motion
  segmentation using event surfaces,'' in \emph{IEEE Conf. Comput. Vis. Pattern
  Recog.}, 2020, pp. 14\,414--14\,423.

\bibitem{7010933}
G.~{Orchard}, C.~{Meyer}, R.~{Etienne-Cummings}, C.~{Posch}, N.~{Thakor}, and
  R.~{Benosman}, ``Hfirst: A temporal approach to object recognition,''
  \emph{IEEE Trans. Pattern Anal. Mach. Intell.}, vol.~37, no.~10, pp.
  2028--2040, Oct 2015.

\bibitem{sironi2018hats}
A.~Sironi, M.~Brambilla, N.~Bourdis, X.~Lagorce, and R.~Benosman, ``Hats:
  Histograms of averaged time surfaces for robust event-based object
  classification,'' in \emph{IEEE Conf. Comput. Vis. Pattern Recog.}, 2018, pp.
  1731--1740.

\bibitem{8946715}
H.~{Rebecq}, R.~{Ranftl}, V.~{Koltun}, and D.~{Scaramuzza}, ``High speed and
  high dynamic range video with an event camera,'' \emph{IEEE Trans. Pattern
  Anal. Mach. Intell.}, pp. 1--1, 2019.

\bibitem{wang2020eventsr}
L.~Wang, T.-K. Kim, and K.-J. Yoon, ``Eventsr: From asynchronous events to
  image reconstruction, restoration, and super-resolution via end-to-end
  adversarial learning,'' in \emph{IEEE Conf. Comput. Vis. Pattern Recog.},
  2020, pp. 8315--8325.

\bibitem{tulyakov2022time}
S.~Tulyakov, A.~Bochicchio, D.~Gehrig, S.~Georgoulis, Y.~Li, and D.~Scaramuzza,
  ``Time lens++: Event-based frame interpolation with parametric non-linear
  flow and multi-scale fusion,'' in \emph{IEEE Conf. Comput. Vis. Pattern
  Recog.}, 2022, pp. 17\,755--17\,764.

\bibitem{zanardi2019cross}
A.~Zanardi, A.~Aumiller, J.~Zilly, A.~Censi, and E.~Frazzoli, ``Cross-modal
  learning filters for rgb-neuromorphic wormhole learning,'' \emph{Robotics:
  Science and System XV}, p. P45, 2019.

\bibitem{Hu_2020_Graft}
Y.~Hu, T.~Delbruck, and S.-C. Liu, ``Learning to exploit multiple vision
  modalities by using grafted networks,'' in \emph{Eur. Conf. Comput. Vis.},
  2020, pp. 85--101.

\bibitem{evkdnet}
Y.~Deng, H.~Chen, H.~Chen, and Y.~Li, ``Learning from images: A distillation
  learning framework for event cameras,'' \emph{IEEE Trans. Image Process.},
  pp. 1--1, 2021.

\bibitem{sun2022ess}
Z.~Sun, N.~Messikommer, D.~Gehrig, and D.~Scaramuzza, ``Ess: Learning
  event-based semantic segmentation from still images,'' in \emph{Eur. Conf.
  Comput. Vis.}\hskip 1em plus 0.5em minus 0.4em\relax Springer, 2022, pp.
  341--357.

\bibitem{messikommer2022bridging}
N.~Messikommer, D.~Gehrig, M.~Gehrig, and D.~Scaramuzza, ``Bridging the gap
  between events and frames through unsupervised domain adaptation,''
  \emph{IEEE Robot. Autom. Lett.}, vol.~7, no.~2, pp. 3515--3522, 2022.

\bibitem{Gehrig_2020_CVPR}
D.~Gehrig, M.~Gehrig, J.~Hidalgo-Carri\'o, and D.~Scaramuzza, ``Video to
  events: Recycling video datasets for event cameras,'' in \emph{IEEE Conf.
  Comput. Vis. Pattern Recog.}, June 2020.

\bibitem{rebecq2018esim}
H.~Rebecq, D.~Gehrig, and D.~Scaramuzza, ``Esim: an open event camera
  simulator,'' in \emph{Conf. on Robot Learn}.\hskip 1em plus 0.5em minus
  0.4em\relax PMLR, 2018, pp. 969--982.

\bibitem{velickovic2018graph}
\BIBentryALTinterwordspacing
P.~Veličković, G.~Cucurull, A.~Casanova, A.~Romero, P.~Liò, and Y.~Bengio,
  ``Graph attention networks,'' in \emph{International Conference on Learning
  Representations}, 2018. [Online]. Available:
  \url{https://openreview.net/forum?id=rJXMpikCZ}
\BIBentrySTDinterwordspacing

\bibitem{8954040}
L.~Wang, Y.~Huang, Y.~Hou, S.~Zhang, and J.~Shan, ``Graph attention convolution
  for point cloud semantic segmentation,'' in \emph{2019 IEEE/CVF Conference on
  Computer Vision and Pattern Recognition (CVPR)}, 2019, pp. 10\,288--10\,297.

\bibitem{redmon2016you}
J.~Redmon, S.~Divvala, R.~Girshick, and A.~Farhadi, ``You only look once:
  Unified, real-time object detection,'' in \emph{IEEE Conf. Comput. Vis.
  Pattern Recog.}, 2016, pp. 779--788.

\bibitem{pix2pix2017}
P.~Isola, J.-Y. Zhu, T.~Zhou, and A.~A. Efros, ``Image-to-image translation
  with conditional adversarial networks,'' in \emph{IEEE Conf. Comput. Vis.
  Pattern Recog.}, 2017.

\bibitem{huang2021spatio}
S.~Huang, Y.~Xie, S.-C. Zhu, and Y.~Zhu, ``Spatio-temporal self-supervised
  representation learning for 3d point clouds,'' in \emph{IEEE Conf. Comput.
  Vis. Pattern Recog.}, 2021, pp. 6535--6545.

\bibitem{orchard2015converting}
G.~Orchard, A.~Jayawant, G.~K. Cohen, and N.~Thakor, ``Converting static image
  datasets to spiking neuromorphic datasets using saccades,'' \emph{Front.
  Neurosci.}, vol.~9, p. 437, 2015.

\bibitem{li2017cifar10}
H.~Li, H.~Liu, X.~Ji, G.~Li, and L.~Shi, ``Cifar10-dvs: An event-stream dataset
  for object classification,'' \emph{Front. Neurosci.}, vol.~11, p. 309, 2017.

\bibitem{amir2017low}
A.~Amir, B.~Taba, D.~Berg, T.~Melano, J.~McKinstry, C.~Di~Nolfo, T.~Nayak,
  A.~Andreopoulos, G.~Garreau, M.~Mendoza \emph{et~al.}, ``A low power, fully
  event-based gesture recognition system,'' in \emph{IEEE Conf. Comput. Vis.
  Pattern Recog.}, 2017, pp. 7243--7252.

\bibitem{he2016deep}
K.~He, X.~Zhang, S.~Ren, and J.~Sun, ``Deep residual learning for image
  recognition,'' in \emph{IEEE Conf. Comput. Vis. Pattern Recog.}, 2016, pp.
  770--778.

\bibitem{chen2021deep}
C.-F.~R. Chen, R.~Panda, K.~Ramakrishnan, R.~Feris, J.~Cohn, A.~Oliva, and
  Q.~Fan, ``Deep analysis of cnn-based spatio-temporal representations for
  action recognition,'' in \emph{Proceedings of the IEEE/CVF Conference on
  Computer Vision and Pattern Recognition}, 2021, pp. 6165--6175.

\bibitem{deng2009imagenet}
J.~Deng, W.~Dong, R.~Socher, L.-J. Li, K.~Li, and L.~Fei-Fei, ``Imagenet: A
  large-scale hierarchical image database,'' in \emph{IEEE Conf. Comput. Vis.
  Pattern Recog.}, 2009, pp. 248--255.

\bibitem{LIAF-Net}
Z.~Wu, H.~Zhang, Y.~Lin, G.~Li, M.~Wang, and Y.~Tang, ``Liaf-net: Leaky
  integrate and analog fire network for lightweight and efficient
  spatiotemporal information processing,'' \emph{IEEE Transactions on Neural
  Networks and Learning Systems}, vol.~33, no.~11, pp. 6249--6262, 2022.

\bibitem{yao2021temporal}
M.~Yao, H.~Gao, G.~Zhao, D.~Wang, Y.~Lin, Z.~Yang, and G.~Li, ``Temporal-wise
  attention spiking neural networks for event streams classification,'' in
  \emph{Int. Conf. Comput. Vis.}, 2021, pp. 10\,221--10\,230.

\bibitem{cannici2019asynchronous}
M.~Cannici, M.~Ciccone, A.~Romanoni, and M.~Matteucci, ``Asynchronous
  convolutional networks for object detection in neuromorphic cameras,'' in
  \emph{IEEE Conf. Comput. Vis. Pattern Recog. Workshops}, 2019, pp. 0--0.

\bibitem{Islam_2021_ICCV}
M.~A. Islam, M.~Kowal, S.~Jia, K.~G. Derpanis, and N.~D.~B. Bruce, ``Global
  pooling, more than meets the eye: Position information is encoded
  channel-wise in cnns,'' in \emph{Int. Conf. Comput. Vis.}, October 2021, pp.
  793--801.

\bibitem{orchard2014bioinspired}
G.~Orchard and R.~Etienne-Cummings, ``Bioinspired visual motion estimation,''
  \emph{Proceedings of the IEEE}, vol. 102, no.~10, pp. 1520--1536, 2014.

\end{thebibliography}

\end{document}